\documentclass{article} 

\usepackage{graphicx}
\usepackage{listings}
\usepackage{enumitem}

\usepackage{color}
\usepackage{xcolor}
\usepackage{xspace}
\usepackage{balance}
\usepackage{subcaption}

\usepackage[utf8]{inputenc} 
\usepackage[T1]{fontenc}    
\usepackage{hyperref}       
\usepackage{url}            
\usepackage{booktabs}       
\usepackage{amsfonts}       
\usepackage{nicefrac}       
\usepackage{ulem}   

\usepackage{ifthen}
\usepackage{setspace}  
\newcommand{\edit}[1]{\textcolor{blue}{#1}}
\newcommand{\redit}[1]{\textcolor{red}{#1}}
\renewcommand{\edit}[1]{{#1}}
\renewcommand{\redit}[1]{{#1}}

\usepackage{amsmath}

\newboolean{techreport}
\setboolean{techreport}{true}
\newcommand{\tech}{\boolean{techreport}}

\ifthenelse{\tech}{
\usepackage[preprint,nonatbib]{neurips2024}
\usepackage{theorem}
\title{\Large \bf \papaya Federated Analytics Stack:\\ Engineering Privacy, Scalability and Practicality}
}
{
\usepackage{usenix-2020-09}
\pagestyle{empty}
\title{\Large \bf \papaya Federated Analytics Stack:\\ Engineering Privacy, Scalability and Practicality}

\author{}
}

\newtheorem{definition}{Definition}
\newcommand{\tightitem}{}
\newcommand{\papaya}{\textsc{Papaya}\xspace}

\newcommand{\orcs}{UO\xspace}
\renewcommand{\paragraph}[1]{\noindent\textbf{#1}}
\renewcommand{\subparagraph}[1]{\smallskip\noindent\textit{#1}}

\newcommand{\COUNT}{\texttt{COUNT}\xspace}
\newcommand{\SUM}{\texttt{SUM}\space}
\newcommand{\eat}[1]{}
\begin{document}

\author{
\edit{\rm Harish Srinivas} \and 
\edit{\rm Graham Cormode} \and 
\edit{\rm Mehrdad Honarkhah} \and 
\edit{\rm Samuel Lurye} \and 
\edit{\rm Jonathan Hehir} \and 
\edit{\rm Lunwen He} \and  
\edit{\rm George Hong} \and  
\edit{\rm Ahmed Magdy} \and 
\edit{\rm Dzmitry Huba} \and 
\edit{\rm Kaikai Wang} \and 
\edit{\rm Shen Guo}\\[3pt] Meta \and 
\edit{\rm Shoubhik Bhattacharya}}
\date{\edit{Meta}}

\ifthenelse{\tech}{
\renewcommand{\papaya}{\textsc{Papaya}\xspace}
}
{
}

\maketitle

\begin{abstract}
Cross-device \textit{Federated Analytics} (FA) is a distributed computation paradigm designed to answer analytics queries about and derive insights from data held locally on users’ devices. On-device computations combined with other privacy and security measures ensure that only minimal data is transmitted off-device, achieving a high standard of data protection. Despite FA’s broad relevance, the applicability of existing FA systems is limited by compromised accuracy; lack of flexibility for data analytics; and an inability to scale effectively. In this paper, we describe our approach to combine privacy, scalability, and practicality to build and deploy a system that overcomes these limitations. 
\ifthenelse{\tech}{Our FA system}{The \papaya FA system}  leverages trusted execution environments (TEEs) and optimizes the use of on-device computing resources to facilitate federated data processing across large fleets of devices, while ensuring robust, defensible, and verifiable privacy safeguards. We focus on federated analytics (statistics and monitoring), in contrast to systems for Federated Learning (ML workloads), and we flag the key  differences.  
\end{abstract}


\section{Introduction}

The scale of distributed systems that allow users to transact and interact have grown massively over the past two decades, prompted by the widespread availability of powerful mobile devices: primarily, smartphones, but also tablets, smart watches, and other wearables. 
These systems are typically mediated by a centralized organization, such as Google, Apple, Meta, Bytedance, or other players broadly in the ``high tech'' arena. 
Providing a positive user experience in such systems relies on their operators monitoring a variety of metrics, and gathering data to inform ML workflows such as recommender systems.  
Consequently, it is necessary to instrument the system for data collection and analysis of user activities and preferences. We refer to this process as \textit{analytics}. 

Increasingly, there are strong motivations to ensure high levels of privacy for analytics.  
Internally, companies recognize the high expectations of users that their data will be used responsibly, and will be protected from threats. 
Externally, a range of laws and regulations govern how organizations must manage data safely, including \redit{GDPR\footnote{\url{https://gdpr-info.eu/}}, ePD\footnote{\url{https://www.edps.europa.eu/data-protection/our-work/}},
and CCPA\footnote{\url{https://oag.ca.gov/privacy/ccpa}}. 
These mean that analytics systems must have strong protections on how user data is stored and processed. 
Meanwhile, we design  the system to operate effectively, so
that monitoring queries can be evaluated accurately and unobtrusively.} 

In response to these needs, the model of federated computation, and specifically federated analytics (FA), has recently risen to prominence~\cite{Bharadwaj:Cormode:22}. 
This model embodies the notion that data gathered by client devices should remain on those devices, and only minimal summary information should be sent from devices to a server, which can compute and release 
aggregate results. 
Beyond this data minimization paradigm, additional protections can be instantiated, to ensure:
deidentification (the identity of the client is dissociated from the messages they send); 
security (data is encrypted to prevent it being observed by unauthorised parties while in transit or at rest); and 
privacy (the results of a query do not disclose information about any individual participant). 

Putting this philosophy into practice requires significant levels of systems and algorithm engineering. 
Careful design choices are needed to ensure that components can cope with the scale of millions to billions of active participants. 
Components must interact reliably, so that the security and privacy guarantees compose constructively, and do not allow leakage or expose vulnerabilities.  
The system must also be flexible enough to allow complex queries to be evaluated over subsets of users, whose ongoing active participation is not guaranteed.

\eat{
\noindent
\textbf{Our Contributions.}
\sout{In this paper, we describe our efforts to combine privacy, scalability, and practicality to build and deploy a federated analytics stack.  
We expose some of the challenges faced and the pragmatic design decisions made to overcome them.  
Our architecture is designed to orchestrate computation across edge devices and trusted execution environments (TEEs). 
We prioritize the simplicity of code running on TEEs, to ensure a reliable, verifiable, and auditable implementation. 
We entrust each device with the responsibility of managing its own data, providing it with full autonomy over the computations it allows and the timing of their execution. Furthermore, we leverage batched processing to amortize resource usage across queries. 
To ensure the platform is practical for analysts, we surface a flexible computational model that is easy to use. This approach enables the rapid development and deployment of new, complex queries, while addressing both scalability and privacy issues. }
\sout{Finally, we present some empirical observations and experiences that help to calibrate the system and provide insights into its performance.} 
}


\subsection{Our Contributions}
Variations of federated analytics technology are utilized across the industry, 
such as by Google for identifying popular items and producing geographic heatmaps~\cite{Bagdasaryan:23,fedhh,rappor}
and by Apple for autocorrect and user experience purposes~\cite{talwar:23,appledp}.
However, the first generation of FA systems faced challenges that tend to limit their wider applicability.
Some systems suffer from low accuracy due to overly restrictive models of privacy, making them less reliable for deriving valuable insights. 
Others lack the flexibility that analysts need to obtain the information necessary for making data-driven decisions. 
Moreover, several early FA solutions do not scale effectively, preventing them from reaching many millions of devices.

In this paper, we present the Federated Analytics stack at Meta that integrates privacy, scale, and practicality in a cohesive architecture.
The FA stack enables efficient computation across three zones: (1) Device, (2) Trusted Environment, and (3) Untrusted Orchestrator. We first provide an overview of the challenges and then our approach to solve them. 

\medskip
\paragraph{Practicality and Accessibility to Analysts.}
Previous FA systems have struggled with limited expressivity, slow query deployment (time from query creation to result), and requiring deep expertise in federated paradigms—barriers to wider adoption. Our approach addresses these challenges as follows:
\begin{itemize}\tightitem
    \item \textbf{Expressive and Flexible Computational Model.} We provide a powerful yet intuitive model combining local (device-level) data transformations with a minimal set of private cross-device aggregation primitives. Analysts can configure federated computations using familiar SQL queries to define local operations, select cross-device aggregation methods, and specify privacy parameters. This balance of expressivity and simplicity meets most analytical needs without requiring deep technical knowledge. See section \ref{sec:fed_query}.
    \item \textbf{Ad-Hoc Query Support and Fast Iteration.} A rapid iteration cycle is crucial for an effective analytics platform. With active query management, periodic device synchronization with the server, and batched computations, our system allows analysts to go from query creation to results within a day. This quick turnaround facilitates agile data exploration and decision-making.
\end{itemize}

\paragraph{Scalability and Resource Consumption.}
\redit{Our system must tackle two aspects: (1) device population---the total number of devices answering a query, which can vary from thousands to billions, and (2) query volume---the number of active queries within the FA system.} Additionally, the processing must be stateful,  fault tolerant, and mindful of resource usage on constrained devices.
See Sections~\ref{sec:scale} and~\ref{sec:failure}.
\begin{itemize}\tightitem
    \item \textbf{One-shot Algorithms.} Unlike Federated Learning, the system is optimized for only one or a very few (constant) rounds of data collection. We express these algorithms with a flexible primitive, namely Secure Sum and thresholding. 
    \redit{
    Furthermore the algorithms are robust to devices dropping out or being unavailable.
    They also avoid any inter-device dependence, and require that client information can be encapsulated in a single message. 
    We give a discussion of FA algorithms for quantiles in Appendix~\ref{sec:algo}.}
    \item \textbf{Batch Processing.} To reduce resource use, \ifthenelse{\tech}{our systen}{\papaya} batches computations on each device. 
    Each device retrieves all relevant queries, executes them and makes one report back, effectively amortising the  cost across metrics. See Section~\ref{sec:scale}
    \item \textbf{Predictable Query Load.} We randomize the sync and reporting schedules of individual devices to distribute data submission over a defined period, controlled by a system parameter, ensuring a manageable and predictable Query Per Second (QPS) to the TEEs. This is in Section \ref{sec:analysisofspeed}
\end{itemize}

It’s important to highlight that \textit{one-shot algorithms} with batched processing were critical to minimise the computational and communication cost on resource constrained devices and maximise inclusion of all device types.

\medskip\paragraph{Privacy and Security} (Section \ref{sec:privacy}).
Federated systems inherently offer a baseline of privacy through data minimization, i.e., keeping raw user data on the device, but our stack aims to provide a higher privacy standard through well-defined mathematical guarantees. Given the diversity of differential privacy models, varying trust levels among actors, and evolving data analysis needs, privacy management becomes complex. We address this by centering privacy guarantees around the user's device. Each user’s device takes responsibility to ensure that their data is handled properly during its processing and the outputs meet the expected privacy standard. 

\begin{itemize}\tightitem
    \item \textbf{Simple Data Handling Off-device.} TEEs provide a secure environment where the device can verify the behaviour; i.e., how data is handled. We have explicitly designed the TEE code to be simple and use-case agnostic, to ensure a reliable, verifiable, and auditable implementation. Concretely, the only role of this environment is to perform Secure Sum across devices, threshold and apply differentially private noise. 
    With SQL data transformation on device and post processing of the anonymized aggregate on untrusted servers we show that this Secure Sum and threshold functionality is sufficient to support a broadly applicable analytics platform.
    \item \textbf{Device Autonomy and Control.} Each device independently decides which queries to execute and when. \redit{It ensures, via remote attestation, that data processing is conducted off-device on a trusted TEE, running authenticated code}. This mechanism guarantees that user data is only processed under secure and verified conditions
    \item \textbf{Support for Different Privacy Modes.} Different privacy models, such as local and distributed differential privacy, offer varying trade-offs between privacy and utility. Recognizing there is no one-size-fits-all solution, our stack supports both. \redit{For each query, the analyst can choose between on-device or TEE-based noise addition and configure privacy parameters accordingly.} Devices validate these parameters before accepting a query, ensuring that only those queries meeting the user-defined privacy standards are processed.
\end{itemize}

\paragraph{Production-Grade System and Evaluation} (Section \ref{sec:study}).
Building on standard concepts in Federated Analytics, we developed a cohesive architecture that addresses the limitations discussed above, with a focus on resource efficiency and fault tolerance (Section \ref{sec:failure}). We successfully implemented this architecture at production scale. Beyond system design, we present empirical results from real-world experiments involving a population of nearly 100 million Android devices. 
\redit{These experiments carefully ensure system and data heterogeneity, measuring key performance metrics such as the time required to iterate over devices and data, and the effects of variables like time of day and device usage patterns.} 
The large-scale experiments demonstrate the system's ability to handle challenges such as unpredictable device availability, diverse network conditions, and maintaining a predictable Query Per Second (QPS) rate to the TEEs. To the best of our knowledge, this is the first large-scale evaluation of a production FA system of this magnitude.

\smallskip
\paragraph{Similarities and Differences with Federated Learning.}
\edit{Both FA and FL instantiate the federated model of computation to protect client data. However, they differ significantly in their workloads, each posing distinct challenges to the system.  
Thus our FA system has fundamental differences in protocol, execution, scalability and hence is implemented separately.
Specifically, FL systems are optimized for ML model training, which entails a fairly homogeneous set of tasks around sending a (large) model to clients, and obtaining (large) updates back from clients in the form of gradient updates.  
This process is repeated many times across small batches of clients (hundreds to thousands).  
By contrast, the FA system is meant for general purpose analytics on all device types. 
It optimizes for computations with fewer rounds of interaction, smaller messages from client to server, and participation from more clients (as many as billions).
While we can use the FA system to run FL workloads, it would not be a good fit, and vice-versa.}

\smallskip
\paragraph{Example use-cases.}
\redit{The system is used for a range of applications in production.  
Some indicative scenarios include:
counting daily and monthly active users of different products, while ensuring that duplicates are not counted repeatedly~\cite{Hehir:Ting:Cormode:23}; 
identifying popular content (heavy hitters) within different geographic regions; 
producing heatmaps of density of activity at differing levels of granularity~\cite{Bagdasaryan:23}; 
tracking the tail of response time distributions to ensure that SLAs are met and to raise warnings (see Appendix~\ref{sec:quantiles}); 
efficient collection of quality statistics (means, counts and variances) on system health metrics that impact user experience~\cite{Cormode:Markov:Srinivas:24}; 
gathering accuracy and calibration metrics on the performance of deployed federated learning systems; 
populating dashboards for adoption of new features in applications; 
and reporting results of federated experiments (A/B testing) on different user interface designs.}

\section{Background: Secure Computation in TEEs}
\label{sec:securecomp}

No matter the approach, federated analytics requires 
client devices to report some information about their data. 
Therefore, in order to provide robust security and privacy guarantees, it is necessary to rely on \textit{secure computation}. 
The goal of secure computation is to evaluate functions on distributed inputs while disclosing only the result of the computation and revealing as little information about the inputs as possible.
There are two feasible approaches for secure computation that are resilient against a curious or adversarial server and which are suited for a federated workload: secure multiparty computation \edit{(MPC)}~\cite{Cramer:Damgard:Nielsen:15} and trusted execution environments (TEEs)~\cite{Shepherd:Markantonakis:24}. While fully homomorphic encryption (FHE) is often suggested as a third alternative, existing FHE techniques are not flexible or performant enough to operate at scale.

\edit{MPC} involves the use of techniques like secret sharing to instantiate partially homomorphic encryption across multiple co-operating servers~\cite{secagg}. 
Meanwhile, TEEs rely on hardware security to ensure that the TEE operates as described and does not leak any information to an external observer. 
\ifthenelse{\tech}{
We adopt TEEs to implement secure computation, with the understanding that we are placing trust in the hardware manufacturer to correctly implement the TEE functionality. }{}
TEEs provide several features for establishing trust that a unit of code has been executed faithfully and privately:

\begin{itemize}\tightitem
    \item \textbf{Confidentiality}: The state of the code’s execution remains secret, unless it explicitly publishes a message.
\item \textbf{Integrity}: The code’s execution cannot be affected by any party, except by the code explicitly receiving a legal input.
\item
\textbf{Measurement/Attestation}: The TEE can prove to a remote party what code (binary) is executing and what its starting state was, ensuring confidentiality and integrity.
\end{itemize}

\ifthenelse{\tech}{
While TEEs offer crucial methods for establishing trust, one needs to be mindful of the following nuances:
\begin{itemize}\tightitem
\item
\textbf{Scalability}: Trusted hardware trades some performance for security guarantees. 
Transferring data into the TEE can be slow and the memory is limited for certain variations of TEEs compared to commodity (non-secure) servers.
\item
\textbf{Verifiability and Auditability}: A key aspect of trust stems from the ability to verify and audit the code that the TEE executes. To improve transparency, code running in TEEs should be as simple as possible. 
\end{itemize}}{}

\edit{Our focus in this paper is on the trust that clients place on the FA system, and in ensuring that raw information does not leak to the analyst or other clients or even the orchestrator. 
We therefore focus on the security and privacy attached to client messages, and providing verification back to clients that their data is handled correctly. 
Handling clients who try to subvert the protocol by manipulating their responses to ``poison'' the results is not in scope for this work---separate mitigations are in place to ensure that clients are only running authorized binaries.}
TEEs enable us to derive anonymized, aggregated insights from distributed datasets while protecting the data of individual clients.
Via attestation,
clients obtain proof of confidentiality and integrity before data ever leaves their devices. 

\smallskip\paragraph{Enforcing Security through Remote Attestation}.
\edit{
In our deployment, we use Intel SGX hardware to provide TEE functionality in the form of \textit{secure enclaves}, similar to the Papaya FL system and others~\cite{papaya,talwar:23,Eichener:24}.
One needs to be mindful about the possible attacks on SGX and apply appropriate mitigations ~\cite{Fei:Yan:Ding:Xie:22}. With client attestation and hosting on public cloud one can limit the attack vectors. 
}
While TEEs provide confidentiality and integrity guarantees, these guarantees are valuable only if a client device can obtain proof that it is sending its data to a valid TEE running a known, trusted binary before the data leaves the device. 
It is possible to do this by using \textit{remote attestation} as follows:

\begin{enumerate}\tightitem
\item 
Before protocol execution, the TEE code is made available for audit along with the hash of the trusted binary.  Ideally, the code is fully open-sourced to allow public scrutiny. 
\item
The TEE generates an \textit{attestation quote} (AQ). This quote is used to cryptographically verify the initial state of the TEE, the hash of the binary running inside the TEE, the hash of public parameters used to initialize the TEE at runtime, and a Diffie-Hellman (DH) key exchange context used to establish a shared secret with the TEE. Under standard operating assumptions, it is not feasible to forge an AQ.
\item 
Upon receiving an AQ the client checks 
(a) the hash of the running trusted binary is the same as the one published with source code,
(b) the public parameters used to initialize the TEE at runtime are valid, and
(c) the DH key exchange context was actually generated by the TEE.
The client aborts if any of these conditions can't be verified.
\item The client uses the DH key exchange context to establish a shared secret with the TEE, and sends its data encrypted.
\end{enumerate}

Via this protocol,
each client ensures it is talking to a legitimate TEE running trusted code with acceptable parameters.

\section{System Design}
\label{sec:components}

\ifthenelse{\tech}{}{
Within a global-scale distributed system, we identify multiple types of participating entities. 
The system is operated for the benefit of many users (ranging in scale from millions to billions).  
Each user makes use of \textit{client devices} (\textit{clients} for short) to interact with the system. 
Each user can operate multiple devices, and one device can be shared among multiple users (switching between accounts). 
For simplicity, we focus on the case where each user uses a primary client device to access the system. 
The client devices communicate with a number of system servers which deliver the service on behalf of the system operator. 
Again, we simplify this to focus on client device interactions with a single device-facing server, which in turn will interact with various \textit{backend servers} on the operator side. 
In order to offer additional privacy guarantees, the system engages additional \textit{trusted aggregators} outside the direct control of the system operator to perform aggregation of client messages.  
These may be sourced from, e.g., a cloud service provider. 
Within this setting, trusted \textit{analysts} seek to find the answer to various queries about client performance, such as measuring client latency and user engagement with new features. 
We seek a solution where client devices retain control over data generated by user actions, sharing only the minimum information necessary to support monitoring queries. 
This paradigm is \textit{federated analytics}, a.k.a. \textit{FA} \cite{Bharadwaj:Cormode:22}.}

\begin{figure}
    \centering
\ifthenelse{\tech}{
\includegraphics[width=0.8\columnwidth]{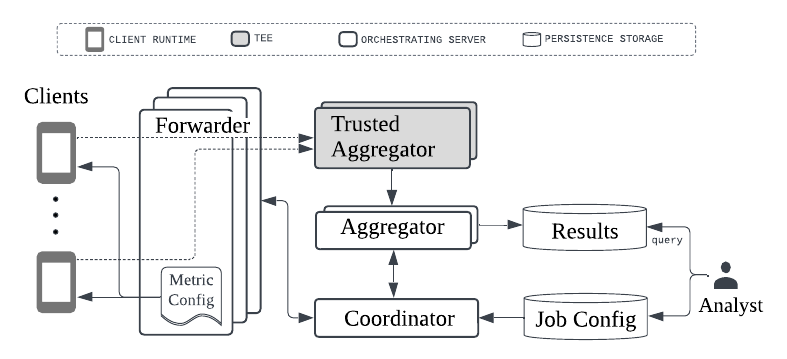}
}{
\includegraphics[width=\columnwidth]{architecture.pdf}

}
    \caption{Overall system architecture}
    \label{fig:arch}
\end{figure}

\ifthenelse{\tech}{Our FA system}{The \papaya FA system} (Figure~\ref{fig:arch}) comprises three applications: an end-user application that runs on client devices, a TEE-based trusted secure aggregator that can run on a system server or a cloud server, and a server application that runs on a data center server and orchestrates the complete protocol.


\subsection{System Overview}
\label{sec:sys_overview}

We introduce the major components of the federated system. 
We describe how an analyst can derive aggregated insights from a distributed dataset in a secure, privacy-preserving way:
\begin{enumerate}\setlength{\itemsep}{0mm}
\item The analyst authors a \textit{federated query}. This has two parts: a SQL-like query that runs on user devices and specifies what data should be uploaded to the server for aggregation; and a server specification that describes how it should aggregate client data  and which privacy techniques to use.
\item The analyst publishes the federated query to the \textit{untrusted orchestrating server} (\orcs), which allocates resources for aggregation and makes the query visible to user devices.
\item The \textit{client runtime} on each user device downloads the federated query spec and retrieves/transforms the relevant data from the on-device data store  \redit{(see Section~\ref{sec:client-runtime})}.
\item The client runtime uses a secure channel to upload data to the \textit{trusted secure aggregator} (TSA) running in a TEE.
\item The TSA uses the federated query spec to aggregate the data across user devices before releasing an anonymized, aggregated result to the orchestrating server.
\item The \orcs uploads the anonymized, aggregated result to a database for consumption by the analyst.
\end{enumerate}
\noindent
We now present each of the above components in more detail.

\subsection{Federated Query}
\label{sec:fed_query}

First, we  introduce some terminology:
\textit{Dimensions} are discrete data attributes representing ``group by'' columns, in the sense that a federated query will group by the specified dimensions before aggregating within each unique set of dimension values. For example, if a query includes \texttt{city} as a dimension, then data points with \texttt{city="Paris"} will be aggregated separately from data points with \texttt{city="New York"}.
%
\textit{Metrics} are quantitative measurements aggregated across clients after grouping by the dimensions of interest, e.g., total time spent on an app across all users.

\smallskip
\noindent
For example, suppose an analyst wants to compute average time spent by city and by day of the week. The  query is split into two parts:

\begin{enumerate}\tightitem
    \item \textit{Data transformation on device}. This is typically defined by a SQL query which runs on the client device. In our example, this local query would retrieve all relevant data on device, group by the dimensions \texttt{city} and \texttt{day}, and then sum the total time spent for each city-day pair. 
    \item \textit{Private aggregation across devices}. This defines how to aggregate the reports from different devices under the chosen privacy-preserving regime. 
    In our example, this will tell the trusted secure aggregator to take the average \texttt{timeSpent} across devices for each city-day pair, and to apply privacy perturbations
    before publishing the result.
\end{enumerate}

\redit{The schemata of on-device tables are present on server and accessible to analysts}. The analyst can configure their query such as in Figure~\ref{config}.
The query result is a table in the data center with one column for each dimension and one column for the metric. 
In our example, there would be columns for \texttt{city}, \texttt{day}, and mean \texttt{timeSpent}. 
Each row would contain a unique city-day pair and the average time spent associated with that pair.

With the above framing, the most common analytical queries can be realized with only a handful of secure aggregation protocols---such as \texttt{COUNT}, \texttt{SUM}, \texttt{MEAN}, and \texttt{QUANTILE}---in combination with on-device local transformation and downstream post-processing to capture use case-specific needs.

\begin{figure}
 \begin{lstlisting}[escapechar=!]
 !\textcolor{teal}{query}!:                         
   !\textcolor{brown}{onDeviceQuery}!: !\textcolor{green!50!black}{"SELECT ..."}!, !\textcolor{black!60}{// SQL to run on device}!
   !\textcolor{brown}{dimensionCols}!: [!\textcolor{green!50!black}{"city", "day"}!] !\textcolor{black!60}{// grouping columns}!       
   !\textcolor{brown}{metricCols}!: !\textcolor{black!60}{// aggregations (e.g., count, mean, ...)}!
     !\textcolor{brown}{mean}!: [!\textcolor{green!50!black}{"timeSpent"}!]
 !\textcolor{teal}{privacy}!:
   !\textcolor{brown}{centralDP}!:
     !\textcolor{brown}{epsilon}!: ...            
   !\textcolor{brown}{kAnonThreshold}!: ...     
 !\textcolor{teal}{output}!: ... !\textcolor{black!60}{// where to persist the anonymized result}\end{lstlisting}
\caption{Example configuration for federated query}
\label{config}
\end{figure}

\subsection{Untrusted Orchestrating Server}
Once the analyst has defined their federated query, it is handed off to the untrusted orchestrating server (\orcs) to begin execution.
The \orcs has the following responsibilities:

\begin{itemize}
    \item Provide centralized coordination to ensure queries progress with enough clients active in the face of hardware failures;
    \item Send active federated queries to clients;
    \item Facilitate communication between clients and TSAs; and
    \item Publish query results to persistent storage.
\end{itemize}
\noindent
The \orcs has several sub-components:

\begin{itemize} \setlength{\itemsep}{0mm}
    \item A fleet of \textit{aggregators}. Each federated query is assigned to a single aggregator at a time. The assigned aggregator is responsible for allocating a TSA for the query, requesting periodic results from the TSA, publishing query results to persistent storage and reporting query progress. Each aggregator may be responsible for multiple queries.
    \item A central \textit{coordinator}, which monitors the state of each federated query, assigns each query to an aggregator and builds the list of active queries to broadcast to clients.
    \item A \textit{forwarder} layer, which handles incoming client requests and forwards them to the relevant backend components.
\end{itemize}

After a federated query is registered with the \orcs, it becomes visible to clients, which retrieve query instructions from the forwarder layer using the client runtime.

\subsection{Client Runtime}\label{sec:client-runtime}
\begin{figure}
    \centering
\ifthenelse{\tech}{
    \includegraphics[width=0.6\columnwidth]{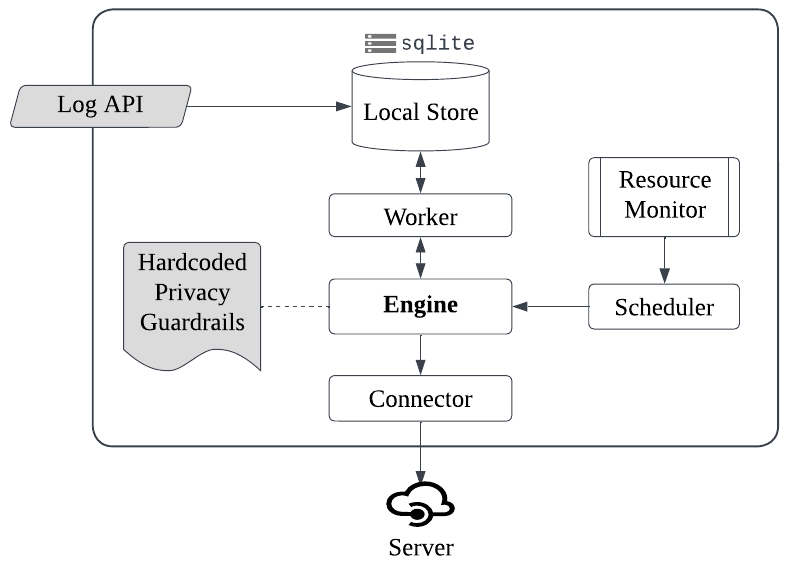}
}{
    \includegraphics[width=0.8\columnwidth]{Client_Runtime.pdf}
}
\caption{Core components of the Client runtime}
    \label{fig:client}
\end{figure}

The client application runs on end-user devices and so needs to be lean, performant and also mindful of the resources consumed by the process. 
The client runtime (Figure~\ref{fig:client}) provides:

\begin{itemize}\tightitem
    \item A \textit{local store} that securely persists data on the device. It manages data lifetime and scope, and provides the ability to run simple analytic functions over the data.
    \item An \textit{engine} to execute the client protocol, described below.
    \item A \textit{scheduler} to monitor the resources consumed and invoke the engine if the device is idle and cumulative resources consumed by the runtime are below a set threshold.
\end{itemize}

\noindent
The client protocol is split into selection and execution phases:

\smallskip\paragraph{Selection Phase}. Each client periodically fetches the list of active federated queries by polling the \orcs (subject to a self-enforced daily limit on total resources consumed). 
Each query configuration in the list has  the instructions the client needs to execute locally for each query. 
Each query configuration may also contain extra information the client uses to decide whether or not it can execute the query. 
\edit{
For example, it may include privacy-related parameters to be used for aggregation on the server side, and the client can reject a query if these parameters do not meet locally enforced guardrails (e.g., by capping the number of queries per day, or barring access to certain features).} 
The query configuration could also include a client subsampling rate, where the client uses its own randomness to reject the query with some probability; this can be leveraged to establish stronger privacy 
guarantees~\cite{talwar:23}. 
Last, the client inspects its local state to see if it has any new data to report for each query. 
After these steps, the client has a list of queries to execute.

\smallskip
\paragraph{Execution Phase}. The goal of the execution phase is for the client to calculate the required outputs for each query from the selection phase, then share the outputs with the TSA assigned to each query. The client splits the set of queries into batches. 
For each task in a batch, the client retrieves and transforms the relevant data; validates the TSA and establishes an encrypted channel via remote attestation; then encrypts its data and sends the encrypted reports to the server, which forwards them to the target TSAs.

\subsection{Trusted Secure Aggregator}
\label{sec:tsa}

The TSA is the component of our system that handles the concrete aggregation logic to combine data from different clients. One TSA corresponds to one federated query, though there can be multiple independent TSA instances on the same physical host.
The TSA runs inside a TEE and uses remote attestation to establish trust and shared secrets with clients. Clients encrypt their data before sending it to the TSA, so it is the \textit{only} component of our backend that sees plaintext data from individual client reports.
Once enough time has passed and enough client devices have reported, the TSA releases an aggregated anonymized result to the \orcs. 

\begin{figure*}
\centering
\ifthenelse{\tech}{
    \includegraphics[width=\textwidth]{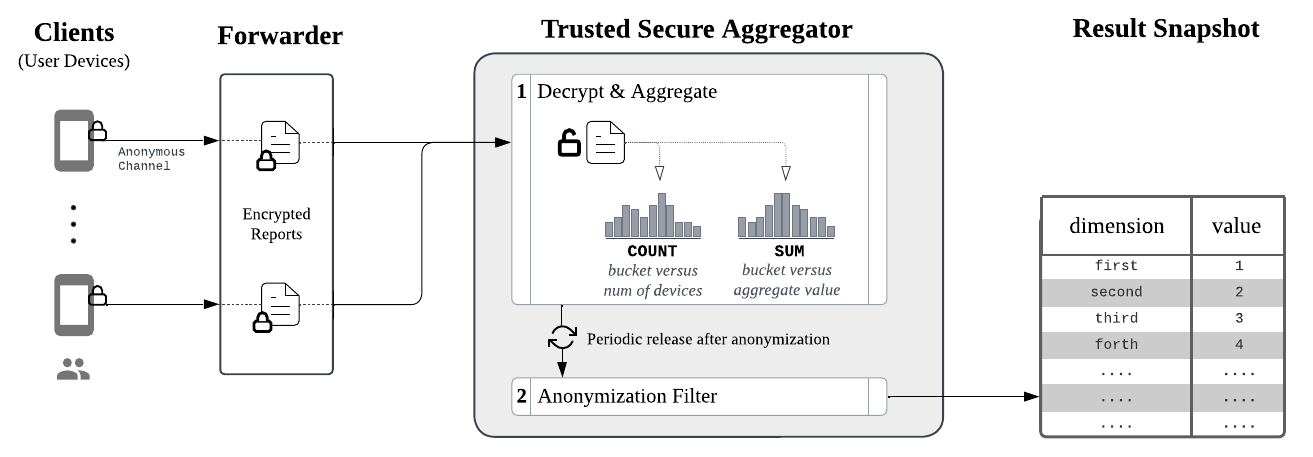}
}{
    \includegraphics[width=0.8\textwidth]{SST.pdf}
}
    \caption{Secure Sum and Thresholding (SST)}
    \label{fig:sst}
\end{figure*}

\paragraph{Secure Sum and Thresholding (SST).}
Many analytical queries can be realized with only a handful of private aggregation protocols such as \texttt{COUNT}, \texttt{SUM}, \texttt{MEAN} and \texttt{QUANTILE} (Section~\ref{sec:fed_query}). 
To further streamline this, our system implements each aggregation primitive via private sparse histograms. Here, \textit{histogram}  refers to taking a set of key-value pairs from distributed client devices and outputting a map from keys (or ``buckets'') to two quantities: the sum of values for the key across all clients with that key, and the count of clients that reported a value for the key.
Histograms are one of the most used queries within FA, and many FA queries rely on histograms as a building block, including prefix queries, range queries, heavy hitters, and quantiles. 
Specifically, these queries use histograms over data with different bucket granularities to build a picture of the data distribution. 
Histograms allow us to address privacy and scalability with a single aggregation primitive.
The TSA running in a TEE allows us to build histograms while maintaining strong privacy and security guarantees via a \textit{Secure Sum and thresholding (SST)} process (Figure~\ref{fig:sst}). 
This process, which can run from minutes to days, goes as follows:

\begin{enumerate}\tightitem
    \item At the start of federated query execution, a TSA is initialized with an empty histogram.
    \item Each client validates the TSA and shares encrypted key-value pairs (essentially a ``mini'' histogram) with the TSA if validation succeeds, using remote attestation (Section~\ref{sec:securecomp}).
    \item The TSA decrypts each client's key-value pairs and immediately aggregates them into the histogram before discarding the individual client data.
    \item Once enough clients have reported and enough time has passed \redit{(see Section~\ref{sec:analysisofspeed})}, the TSA anonymizes the histogram by applying privacy noise to both the sum value and client count value for each bucket. 
    It filters out any buckets with a noisy client count below a threshold specified by the analyst---i.e., if too few clients report a value for some bucket, that bucket will not be released from the TSA (see Section~\ref{sec:hist}).
    \item The anonymized histogram is released to the \orcs.
\end{enumerate}

\subsection{Handling Scale}
\label{sec:scale}
{ 
For scalability, we employ strategies that address both the volume of devices and the number of queries, 
while enabling random and unbiased device sampling.}

\paragraph{Single-Query Scalability}. Employing one-shot algorithms built on Secure Sum and thresholding, each device submits data only once per query, reducing traffic and processing load. 
To mitigate the ``thundering herd'' problem, where multiple devices attempt to connect simultaneously, we randomize the reporting schedules of individual clients.  
This distributes the data submission over a defined period, controlled by a system parameter, ensuring a manageable and predictable QPS to the TEEs (see Section~\ref{sec:study} and Figure~\ref{fig:coverage}). 
The server immediately aggregates incoming reports, using memory proportional to the histogram's dimensions and simplifying the processing involved. 
Our experiments show a single server is sufficient for one query, but this can be expanded to a tree-level aggregation scheme to distribute the workload.

\paragraph{Multi-Query Scalability}. To handle multiple queries, our system batches computations on each device and shards by query on the server. Devices receive all relevant metrics at once, execute their computations, and report back collectively. 
This setup amortizes process initiation costs and server communication, scaling with query volume.

\subsection{Handling Failures}
\label{sec:failure}
There are multiple potential failure modes, which we classify into two clusters: faulty clients, and faulty servers. 
For simplicity, we consider faulty connections and clients together, 
since clients often have unreliable connections that are subject to interruptions. 
\redit{To circumvent this, we break the device work into batches of size $\sim10$. 
This size was empirically determined and can be fine tuned on the respective platform to make sure interruptions are infrequent.}
If interruptions do happen, we retry during the next period. 
The client computation is idempotent and will occur again until a successful acknowledgement (\texttt{ACK}) for that metric has been received from the server.

This leaves failures that occur on the server side, especially the aggregator and TSA, as they are stateful. 
\edit{Each aggregator-TSA pair stores periodic snapshots of query progress (every few minutes) that would allow a different aggregator-TSA pair to recover intermediate results and resume query execution in case of failure.
The coordinator component of the \orcs can detect fatal query execution errors and will reassign and restart a query on a new aggregator when this occurs. 
If the coordinator itself fails, a new coordinator instance is started, recovering the previous state  from persistent storage.}

An important issue here is the privacy of the intermediate aggregation state with partial query results. If the partial results meet the privacy requirements as defined in the query, they can be released and stored in plaintext. Alternatively, the intermediate results can be stored in an encrypted form that is only accessible by another TEE running the same binary as the TEE that generated the results. This is achieved by maintaining a separate group of TEEs responsible for generating, storing and replicating encryption keys. 
Maintaining robust privacy guarantees is possible with careful bookkeeping and replication logic~\cite{elephantdp}. Encrypted aggregation state becomes unrecoverable when the associated encryption key is lost, which occurs if and only if a majority of the TEEs with that key fail. This is not a significant obstacle in practice. Because we replicate encryption keys rather than actual data, it is relatively inexpensive to increase the number of replication nodes. 
As intermediate aggregation state is cumulative, we only need the latest encrypted data. 
Increasing the frequency of encrypted snapshot generation further reduces the impact of data loss.


A final consideration for the robustness comes from the federated algorithms themselves. 
In Appendix~\ref{sec:algo}, we require that the algorithms do not need participation from all clients, expecting that drop outs will occur as a matter of course. 
\edit{
Additionally, if a client is malicious and tries to `poison' its output the effects are negligible due to (1) its contribution is bounded per report on the TEE prior to aggregation, (2) the typical large scale of operation. }

\section{Privacy and Security}
\label{sec:privacy}
\ifthenelse{\tech}{
In Section~\ref{sec:components}, we shared how the stack is able to capture complex queries at scale via secure aggregation of histograms. 
This provides an intuitive level of data protection by insisting that clients do not share their full information and all of the client reports are securely aggregated only to reveal aggregated statistics. 
In this section, we discuss how data is further protected as it is handled by the federated analytics stack. }{
Secure Aggregation provides a baseline of data protection by insisting that clients do not share raw information, and only aggregated statistics are revealed. 
We now discuss how data is further protected as it is handled by the FA stack
via authentication, access control, and explicit privacy noise addition. 
}

\subsection{Secure Data Handling}
\paragraph{Storage.}
Our FA system protects data at rest by design by storing the data on the user's device. 
The device application manages data scope and lifetime, with encryption and access controls applied to protect data from potential adversaries. \redit{Data retention time is configurable with max lifetime (typically 30 days) hard-coded in the application as a guardrail.} 

\paragraph{Communication.}
Protecting data during transit is a well-studied problem, and we rely on the standard Transport Layer Security (TLS) protocol to secure network communications. 
To further minimize leakage of data, 
communications happen via anonymous authenticated channels, making use of the Anonymous Credentials Service (ACS) library \cite{acs,dit}. 
Thus, the platform is unaware of the identity of the client. 

\paragraph{Processing.}
Secure processing is nuanced due to intricate interactions. 
Each user's device takes responsibility to ensure that their data is handled properly during its processing. 
All data from one user is either handled on their  device or in an environment where their device can verify the data handling.
%


\smallskip
\noindent
\textbf{Device control over computation.}
A first component of privacy is admission control. Each device determines which computations to run and when, based on eligibility criteria like previous FA participation, geographic region, hardware type, software version, user features, available data, privacy guardrails, and local randomness. In this way, client devices have total control over their participation in the FA process.


\smallskip
\noindent
\textbf{Validation before sharing.} Because additional data processing occurs on the TSA server, it is not sufficient that the device controls its own computation.
The device must also ensure that the server will handle its data properly and privately. In order for the device to make a fully informed decision about whether to provide data without trusting the \orcs, the device can use remote attestation as described in Section~\ref{sec:securecomp}. 
With this protocol, before sharing any data, the client validates that its data will be processed by a legitimate TEE running a known binary. We have explicitly designed the TEE code to be simple and use case-agnostic, ensuring a reliable, verifiable, and auditable implementation. 
Thus, prior to any communication, clients  obtain proof that their data will be handled as expected.

\subsection{Private Handling of Outputs}
\label{sec:hist}


The previous section discussed how the system provides protections for the data while it is  processed in order to compute aggregate statistics. 
However, aggregating statistics alone is not necessarily sufficient to preserve a strong notion of privacy, as some statistics can still inform on the state of the client.  
We enforce additional measures in order to meet the higher bar of differential privacy. 
%

\begin{definition}[\cite{Dwork:Roth:14}]
\label{def:approx-dp}
Approximate differential privacy (DP) is parameterized by $(\epsilon, \delta)$ and requires that we define a randomized mechanism $M$ such that,
for pairs of inputs $X$, $X'$ that are neighboring (differing in the presence of a single piece of information) and for any possible output  set $O$, $\Pr[M(X) \in O] \leq \exp(\epsilon) \Pr[M(X') \in O] + \delta$. 
\end{definition}

%
Many randomized DP mechanisms have been defined for different applications; here, we focus on obtaining privacy for a histogram of values, where adding zero-mean Gaussian noise 
to each count  gives $(\epsilon, \delta)$-DP~\cite{Dwork:Roth:14}. 

After adding noise, we apply $k$-anonymity, where any counts below $k$ are removed from reports. 
\edit{While $k$-anonymity on its own has been deprecated as a formal privacy tool, this additional step provides an intuitive notion of privacy to our histograms for both users and decision makers---in contrast with the formal DP guarantee, which requires some statistical sophistication to appreciate.
Moreover, when histogram dimensions are not known \textit{a priori}, this thresholding step is critical to the DP guarantee~\cite{wilkins2024exact}.}

We briefly show DP models
which vary in how the notion of neighboring inputs
is defined and at what granularity the definition is enforced: centrally or by each client individually.

\label{sec:histogram-privacy}

\paragraph{Central DP at the Enclave.}
Applying DP via the central aggregator (TEE) is the most straightforward way to obtain DP. 
The TEE can compute the exact histogram, in the form of a \SUM or \COUNT aggregation,  
then add noise to each value in the bucket of the histogram to achieve differential privacy. 

\paragraph{Local DP.}
Local differential privacy takes the standard DP definition, but applies it to every message from every device.
For \COUNT-queries we can represent the user's input as a 1-hot vector and randomly flip the bits, or pick a value to report from an appropriate probability distribution.  
The enclave or server aggregates the reports from all devices, and performs a statistical de-biasing step to obtain the estimated histogram.

\paragraph{Distributed Privacy Noise.}
In the distributed DP model,
each client adds a small amount of noise, so that after aggregation the noise is sufficient to give the desired DP guarantee.
Each client builds a ``mini histogram'' 
with the full set of buckets, records its values, and draws random noise to add.  
The noisy mini histograms are sent to the TSA, which sums them and checks that enough have been received to achieve the central noise requirements. 
It then performs $k$-anonymity enforcement and releases the final result.
We use the ``sample-and-threshold'' approach to distributed noise addition, where the uncertainty is introduced due to client randomly deciding whether or not to participate in the data collection~\cite{Bharadwaj:Cormode:22}.  

\paragraph{Periodic Data Release.}
An important consideration in anonymizing the histogram is when results should be released.
Client devices have intermittent connectivity and availability; there can be a delay on the order of days before enough participate in FA query to report results with confidence. 
This poses a challenge to our system, \edit{as analysts want results as soon as possible}. 
We balance the desire to give up-to-date results against the privacy cost of making multiple disclosures. 
The TSA sends partial results to the orchestrating server every few hours. 
To uphold the privacy guarantee, 
we limit the number of times the TEE releases partial results.
The overall DP privacy parameters ($\epsilon, \delta$) set by the query configuration are budgeted across all releases, using ``composition'' results~\cite{Dwork:Roth:14}.


\begin{figure}[t]
    \centering
    \begin{subfigure}{0.5\columnwidth}\centering
      \includegraphics[width=1.0\textwidth]{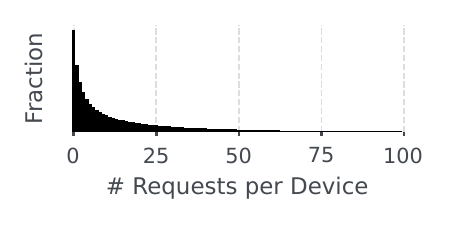}
      \caption{Daily values stored per device}
      \label{fig:reports}
    \end{subfigure}%
    \begin{subfigure}{0.5\columnwidth}\centering
      \includegraphics[width=1.0\textwidth]{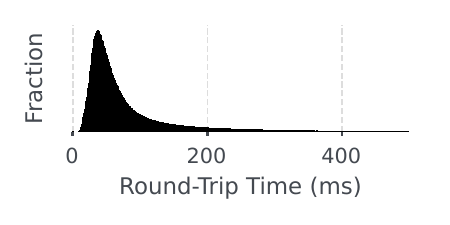}
      \caption{Round Trip Times (RTT)}  
      \label{fig:rtts}
    \end{subfigure}
    \caption{Heterogeneity of data: \textit{(a)} number of sampled requests per device, \textit{(b)} round-trip times}
    \label{fig:heterogeneity}
\end{figure}

\section{Empirical Study}
\label{sec:study}

\ifthenelse{\tech}{To validate and measure our system with high fidelity, we developed our production benchmark to carefully emulate typical challenges from real-world scenarios. 
It is representative of data and system heterogeneity, and is used to measure iteration speed, accuracy, and privacy cost.
\subsection{Experiment Setup}}

We perform our experimental study using queries in our production deployment of \ifthenelse{\tech}{the FA stack}{the \papaya FA stack}.  
Queries are issued on various system health metrics to ensure smooth running of different systems. 
Here, we focus on queries of network activity: round trip times (RTT) per network request and request volume. 
These queries give insights into system and device heterogeneity.
\redit{They are used to track system performance and are an important canary for regression issues.}

Formally, each device $i$ records a set of $n_i$ round trip times $x_{i1}, \dots, x_{i n_i}$, which are aggregated on-device into a local RTT histogram $u_i = (u_{i1}, \dots, u_{iB})$, where $u_{ik}$ denotes the number of RTT values from device $i$ assigned to bucket $k$. 
We compute a federated histogram $v = \sum_i u_i$, making use of the trusted secure aggregator (TSA, Section~\ref{sec:tsa}). 
To objectively measure performance, the data points $x_{ij}$ are also stored in a central  database (for evaluation purposes only), from which we compute a ground-truth histogram $w = (w_1, \dots, w_B)$, where $w_k$ denotes the total number of data points in bucket $k$ across all devices.
%
For the device activity histogram, 
each device has a single data point of interest, $n_i$ (as defined above). 
\redit{So the local histogram produced is a one-hot vector $u_i \in \{0, 1\}^B$, i.e., $u_i[j]=1$  encodes $n_i$ is in bucket $j$ out of $B$.} 
The rest of the federated histogram collection proceeds as in the RTT case.

Each device has a periodic job to poll the server, compute the local histogram $u_i$, and report to the TSA. 
This job has a 10-second timeout, runs in the background, and is run at most twice per day 
to minimize the burden placed on client devices.
Each device also adds individual randomness on when to initiate reporting, to smooth out traffic load.

Queries run on a population of nearly 100 million Android phones that report the relevant data to Intel SGX devices operated by a secure cloud service provider, under the guidance of the \orcs. 
Data are gathered at both daily and hourly granularities. The number of data points in the hourly reports is proportionately lower than in daily reports. 

Figure~\ref{fig:heterogeneity}
shows the distribution of device data, highlighting two different aspects of heterogeneity. 
Figure~\ref{fig:reports}
shows that usage patterns differ widely across devices, leading to varying dataset sizes on each device. 
While the most common case is for clients to have just a single sampled value to report, it is not unusual for them to have tens, with a few having in excess of 100 values to report. 
Figure~\ref{fig:rtts}
shows the differences in network RTT. 
The mode is around 50~ms RTT, but the distribution stretches out to half a second or more. 
This reflects variation in device resources and capabilities.

\begin{figure}[t]
\centering
\ifthenelse{\tech}{
\begin{subfigure}{0.45\textwidth}
       \includegraphics[width=\textwidth]{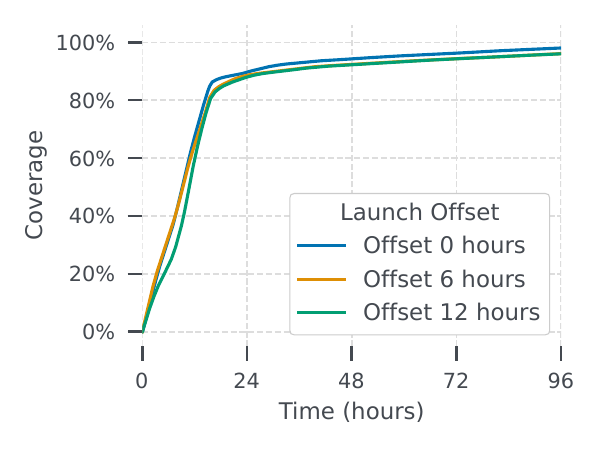}
       \caption{Three query runs}
       \label{fig:coverage-time}
   \end{subfigure}
   \begin{subfigure}{0.45\textwidth}
       \includegraphics[width=\textwidth]{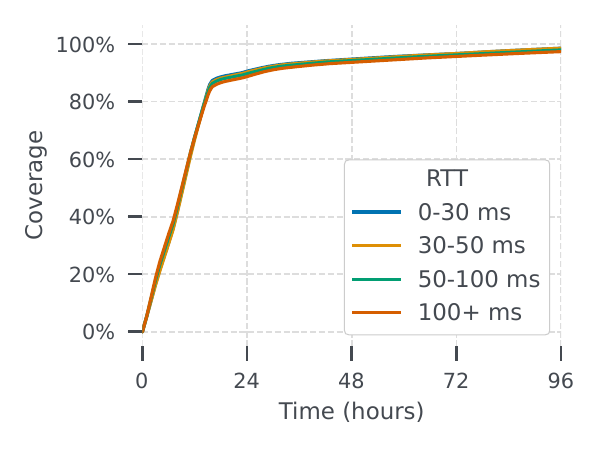}
       \caption{Coverage by RTT}
       \label{fig:coverage-time-rtt}
   \end{subfigure} 

}{
   \begin{subfigure}{0.25\textwidth}
       \includegraphics[width=\textwidth]{plots/coveragevtime.pdf}
       \caption{Three query runs}
       \label{fig:coverage-time}
   \end{subfigure}%
   \begin{subfigure}{0.25\textwidth}
       \includegraphics[width=\textwidth]{plots/coveragerttvtime.pdf}
       \caption{Coverage by RTT}
       \label{fig:coverage-time-rtt}
   \end{subfigure} 
}
\caption{Coverage of the device population over time: \textit{(a}) for different query execution times, \textit{(b)} by device latency}
\label{fig:coverage}
\end{figure}

\begin{figure*}[t]
\centering
\ifthenelse{\tech}{
    \begin{subfigure}{0.45\textwidth}
        \centering
        \includegraphics[width=1.0\textwidth]{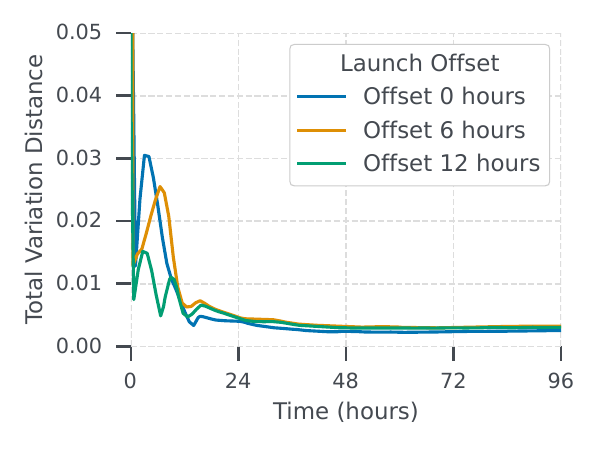}
        \caption{TVD: different offsets}
        \label{fig:tvd-offsets}
    \end{subfigure}%
    \begin{subfigure}{0.45\textwidth}
        \centering
        \includegraphics[width=1.0\textwidth]{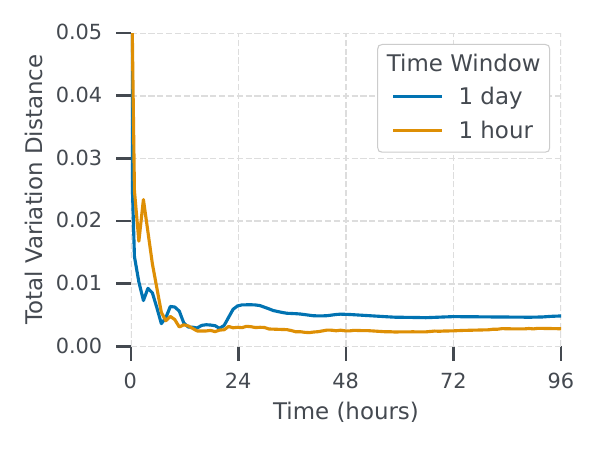}
        \caption{TVD: Different queries}
        \label{fig:tvd-events}
    \end{subfigure}
}
{
    \begin{subfigure}{0.33\textwidth}
        \centering
        \includegraphics[width=1.0\textwidth]{plots/tvdvtime.pdf}
        \caption{TVD: different offsets}
        \label{fig:tvd-offsets}
    \end{subfigure}%
    \begin{subfigure}{0.33\textwidth}
        \centering
        \includegraphics[width=1.0\textwidth]{plots/tvdvtime_numevents.pdf}
        \caption{TVD: Different queries}
        \label{fig:tvd-events}
    \end{subfigure}
    }
    \caption{Accuracy over time: \textit{(a)} for different query execution times, \textit{(b)} for two different queries}
    \label{fig:accuracy}
\end{figure*}

\subsection{Collection Speed and Scalability}
\label{sec:analysisofspeed}

To begin, we evaluate how long it takes for a task to reach all devices and iterate over the data in its entirety. 
It is important to note that in practice a random sample of a fraction of the devices or data is considered sufficient to answer most metrics. 
However, in order to build a thorough understanding of our system, we evaluate the time needed to access all data and devices. 
We measure wall-clock time from the launch of the query  and compute \textit{coverage}, the number of data points processed by the FA task divided by the number of ground-truth points in the database. 
There are many variables that can impact the collection speed, including
time of day (device usage and connectivity patterns vary throughout the day), job restrictions (configuration indicating when a device should not report, based on conditions like network metering, battery status, cumulative quotas, etc.), and other device characteristics (e.g., capabilities, frequency of use, connection quality).

%
%
Figure~\ref{fig:coverage} shows how quickly our system accesses the population over different query executions. 
These queries request the round trip time (RTT) values from devices.
Figure~\ref{fig:coverage-time}
shows three executions of the same query launched at six-hour intervals. 
\edit{
Encouragingly, results are not much influenced by time of day, and so are similar across all runs.}

In fact, the behavior we see in Figure~\ref{fig:coverage} is most strongly influenced by our choice of system parameters: 
We observe a linear growth in coverage up to around 85\% over the initial 16 hours after issuing the query. 
This reflects that clients check into the server at random, with a uniform delay of 14-16 hours.  
Coupled with random offsets, this means that active client reports are spread uniformly through the initial 16 hour period after the query is issued. 
This accounts for a majority of clients. 
The remaining 15\% of less active clients check in more gradually, hitting 90\% in 24 hours and reaching over 96\% coverage after 96 hours, i.e., 4 days from when the query is issued.
We should not expect to attain 100\% coverage, as a small minority of devices may go fully offline, have local storage reset, or otherwise be unable to report in. 
For faster collection, we could narrow the check-in window from 16 hours to a lower value.  
This would speed up the time to reach 85\% coverage, but we expect that the ``long tail'' of less active clients would still require up to a week to check in, due to their sporadic availability.

Figure~\ref{fig:coverage-time-rtt}
tests whether the speed of coverage is correlated with network connection quality, by measuring coverage from a single query based on subsets of the histogram corresponding to different ranges of RTT values. 
The correlations observed are very small, with the coverage rate for low and high latencies following remarkably similar curves.
On close inspection, there is a small effect: low latencies have higher coverage than high latencies. 
This is easiest to see at 16 hours---as time passes, this gap shrinks further. 

Through our experiments, we observed that the majority of resource consumption on devices is driven by process initiation and communication with the server, while the actual computation of metrics is comparatively insignificant. 
This aligns with our hypothesis and design, as unlike Federated Learning, we only execute lightweight SQL queries on the device. Our batched processing setup effectively amortizes these initiation and communication costs, enabling the system to handle many concurrent queries (around 100) efficiently. 



\subsection{Analysis of Accuracy}
\label{sec:analysisofaccuracy}
Next we evaluate the accuracy of a metric computed in the federated setting against the same metric computed from the ground-truth dataset. 
As explained in Section~\ref{sec:fed_query}, all metrics can be computed through data manipulation on the device followed by a handful of aggregation primitives. 
Since each aggregation primitive is built via the histogram operator using the Secure Sum and Threshold primitive (Section~\ref{sec:tsa}), measuring the accuracy of a histogram can be used as a proxy for measuring accuracy of any standard metric.

Letting $n_v = \sum_k v_k$ denote the number of data points in the histogram $v$, the normalized histogram $\bar v = \frac 1{n_v} v$ reflects the relative frequencies per bucket. 
To plot error between federated histogram $v$ and the ground-truth $w$, we use total variation distance (TVD) between their normalized versions, 
\centerline{$
\textstyle
d_\text{TV}(\bar v, \bar w) = \frac 12 \| \bar v - \bar w \|_1 = \max_{S \subseteq \{1, \dots, B\}} \left| \sum_{k \in S} (\bar v_k - \bar w_k) \right|.$}
This straightforward measurement of error reflects the bucket-wise absolute differences between the frequencies observed in 
$\bar v$ vs. $\bar w$ for any subset of histogram buckets.

Figure~\ref{fig:accuracy} shows the total variation distance across several queries as a function of the time since query launch. 
In Figure~\ref{fig:tvd-offsets},
histograms of round trip time (RTT) are created with $B=51$ buckets: 0-10~ms, 10-20~ms, $\dots$, 490-500~ms, 500+~ms. These are created with the same time offsets as in Figure~\ref{fig:coverage}. 
In Figure~\ref{fig:tvd-events},
histograms of device request counts are created at the daily and hourly grain, using $B = 50$ and $B = 15$ buckets (respectively), corresponding to sampled counts of 1, 2, $\dots$, $B-1$, $B$+.
In both cases, we see that it is possible to achieve a very accurate representation: the final TVD is negligible, meaning that the distribution obtained via federated reporting is nearly identical to the ground truth.  
This is obtained quickly: within 12 hours, an accurate result is found, corresponding to when half of clients have checked in (per Figure~\ref{fig:coverage}).  
Even within a few hours, the result is pretty accurate. 


\begin{figure*}[t]
\ifthenelse{\tech}{
\centering
\begin{subfigure}{0.45\textwidth}
        \includegraphics[width=\textwidth]{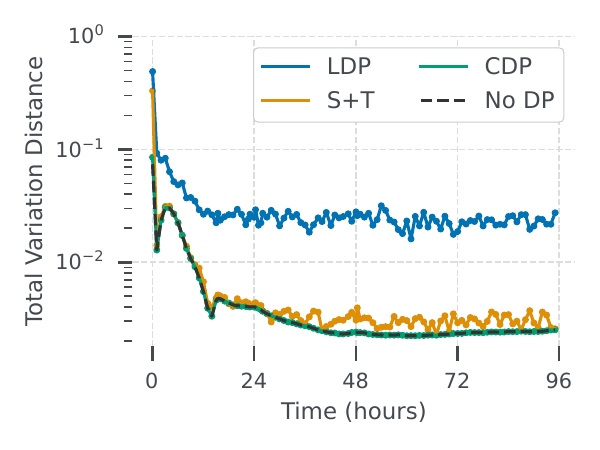}
        \caption{RTT histogram accuracy}
        \label{fig:rtt-privacy-time}
    \end{subfigure}%
    \begin{subfigure}{0.45\textwidth}
        \includegraphics[width=\textwidth]{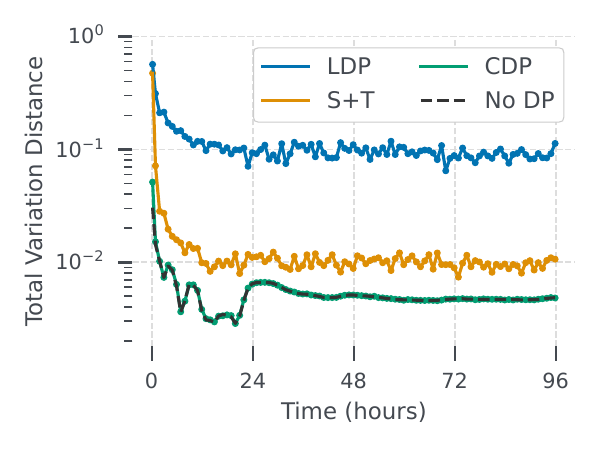}
        \caption{Daily event count histogram accuracy}
        \label{fig:daily-privacy-time}
    \end{subfigure}
    \begin{subfigure}{0.45\textwidth}
        \includegraphics[width=\textwidth]{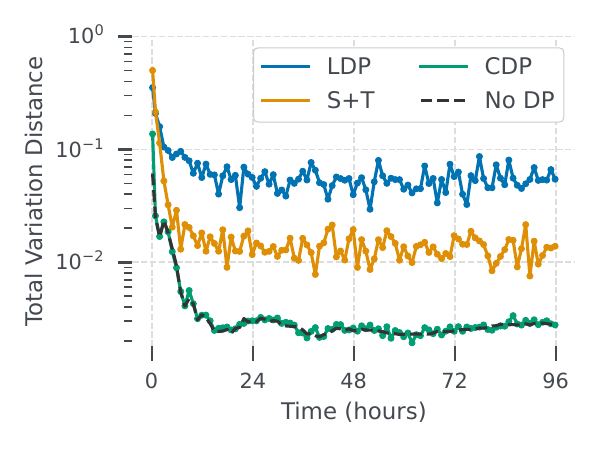}
        \caption{Hourly event count histogram accuracy}
        \label{fig:hourly-privacy-time}
    \end{subfigure}
}{\begin{subfigure}{0.32\textwidth}
        \includegraphics[width=\textwidth]{plots/50_buckets_privacy_time.pdf}
        \caption{RTT histogram accuracy}
        \label{fig:rtt-privacy-time}
    \end{subfigure}
    \begin{subfigure}{0.32\textwidth}
        \includegraphics[width=\textwidth]{plots/1day_num_events_privacy_time.pdf}
        \caption{Daily event count histogram accuracy}
        \label{fig:daily-privacy-time}
    \end{subfigure}
    \begin{subfigure}{0.32\textwidth}
        \includegraphics[width=\textwidth]{plots/1hr_num_events_time.pdf}
        \caption{Hourly event count histogram accuracy}
        \label{fig:hourly-privacy-time}
    \end{subfigure}
}\caption{Experiments on histogram generation with different models of privacy noise addition}
    \label{fig:private-histograms}
\end{figure*}

\subsection{Analysis of Privacy Mechanisms}
\edit{To better guide how different privacy guarantees may affect query accuracy,   
we study how adding differentially private noise impacts the accuracy of histograms.} 
In Section \ref{sec:histogram-privacy}, we described ways to provide a differential privacy guarantee, based on how and where noise is added in the system: centrally at the secure enclave (central DP, \textit{CDP}), by each client locally (local DP, \textit{LDP}), or in a distributed fashion (sample-and-threshold, \textit{S+T}). 
We compare generating histograms in our system under these three different models of privacy via TVD to the ground truth. 
Each data release from the CDP and S+T privacy mechanisms satisfies $(\epsilon, \delta)$-DP (Definition~\ref{def:approx-dp}), and each LDP release satisfies $(\epsilon, 0)$-DP, all with $\epsilon=1, \delta=10^{-8}$.

Recall that against the ground truth, the steady-state total variation distance (Section~\ref{sec:analysisofaccuracy}) is considerably below $10^{-2} = 0.01$ (Figure~\ref{fig:accuracy}). 
Figure~\ref{fig:private-histograms} shows that there is a clear difference in behavior based on the scale of the input population and the privacy model. 
When building the histogram of round trip times (Figure~\ref{fig:rtt-privacy-time}), we see that
local differential privacy (LDP) achieves an order of magnitude more noise than other methods, and this gap does not decay over time. 
\edit{
Nevertheless, this error is in the region of 0.02, which is tolerable for most of our applications.}  
The gap between privacy mechanisms is replicated across other tests (Figures~\ref{fig:daily-privacy-time}, \ref{fig:hourly-privacy-time}), where the error for CDP and S+T is much lower than LDP---for CDP, the result is essentially indistinguishable from the un-noised data collected.  
Results in Figures~\ref{fig:daily-privacy-time} and \ref{fig:hourly-privacy-time} consider the accuracy of reporting histograms on the amount of activity recorded on a daily and hourly basis (respectively). 
Note that the hourly activity counts are proportionately lower than the daily ones.
Here, the hourly activity was 34 times lower than the daily activity, as the measurements correspond to a quieter than average period. 
Hence, the impact of privacy noise introduces more variation for the hourly event counts. 
This is particularly notable for S+T, where thresholding the sampled data leads to more signal loss. 
Meanwhile, error introduced by CDP noise addition remains almost imperceptible on these plots.  

Thus, for histogram generation, privacy can be achieved with minimal impact on accuracy in FA.  
The three privacy techniques shown are arguably incomparable, since they exercise different trust models.  
But even our ``weakest'' model, CDP, is stronger than simply gathering the data on a regular server, since we use secure TEEs to perform the aggregation and noise addition.  
Furthermore, the local model of privacy, as previously deployed at scale~\cite{appledp, rappor}, has the biggest impact on accuracy. 
So, where permissible, we advocate for central or distributed privacy noise within FA systems. 

\redit{Additional experiments on the impact of privacy noise of the accuracy for the quantiles (CDF) query are presented and discussed in Appendix~\ref{sec:quantiles}.}

\section{Related Work}

Instrumenting large scale deployments of distributed clients has been studied widely in the distributed systems and networking communities.  However, building systems with data privacy as a main goal has enjoyed less focus until recently.  

The notion of federated computation has emerged as a distributed paradigm that emphasizes the privacy of the participants~\cite{Bharadwaj:Cormode:22}. 
Systems implementing federated \textit{learning} (FL) have been described by \redit{Google~\cite{BonawitzEGHIIKK19}}, Apple~\cite{applefl}, \redit{LinkedIn~\cite{flint23}}, Snap~\cite{snapfl}, Meta~\cite{papaya} and others in the big tech arena.  
FL is primarily concerned with the training of machine learning models, with examples held by the clients used as the training set. 
Typically, FL workflows have access to a very large number of clients (millions or billions), but engage only a few thousand in each round.  
In each round, a batch of clients suggest improvements to the current model. 
This process is repeated over a large number of rounds, often several thousands in total \cite{googlefl}.  
The duration can thus be days to weeks, allowing for clients that may be slow to respond or only intermittently available.  
Here, 
we focus on analytics tasks that are qualitatively different in nature.  
Rather than concentrate on a single computation task (model training), we seek a system with greater flexibility for processing a range of possible queries.  
These queries are typically processed in few rounds: ideally, a single round or two.  
However, we seek to engage many more clients in each round (at least tens of thousands, often millions), and so the latency to complete a round is higher, due to greater variability in client availability and dropouts.  

This approach to FA was promoted by Google \cite{googlefablog}, with an emphasis on gathering information on frequency distributions.  
The antecedents of this model can be seen in Google’s prior RAPPOR system~\cite{rappor} and Apple’s differential privacy deployment~\cite{appledp}.  Both these systems use  local differential privacy to mask client inputs, building on the classical technique of randomized response~\cite{warner65}.  
Clients are prompted to provide (perturbed) information on browsing and typing behavior (e.g., frequently visited webpages, new words), in order to improve autocomplete suggestions and detect malware.  
This information is gathered regularly and perturbed based on a sparse vector encoding with random noise.  
These first systems demonstrated the utility of private information gathering.  
However, the level of noise due to local DP is high, requiring a very large population of clients and a loose setting of privacy parameters to get  actionable information~\cite{appleprivacyloss}.

Subsequent developments have sought to broaden the range of statistics that can be collected, and to provide stronger privacy and security protections.
Work on secure aggregation~\cite{secagg, bellsecureaggregation} highlights the power of a primitive that computes the sum of inputs from a large number of clients without revealing any intermediate values.  
Google’s described solution uses interactions between clients to recover from other clients dropping out during the protocol, via cryptographic techniques (threshold secret sharing) and a single coordinating server.  
Other approaches to implement secure aggregation achieve a similar functionality but remove client-client communications by instead having two or more independent aggregating servers.
PRIO, implemented by Mozilla, performs aggregation of inputs from clients based on secret sharing, along with simple zero-knowledge proofs to check that client inputs are well-formed~\cite{PRIO, Poplar, Mozilla}.  
Apple’s sampling-based system adopts  this model, where two servers (\textit{leader} and \textit{helper}, as in~\cite{ietf-ppm-dap-08}) interact to perform the aggregation~\cite{talwar:23}. 
\edit{Systems such as Honeycrisp and Orchard use additively homomorphic cryptography to aggregate client reports across multiple servers~\cite{honeycrisp,RothZHP20}.  
For our implementation, we adopt TEEs for aggregation, to simplify key management and scaling issues, and to be able to handle the thresholding primitive.}

Most effort in FA has been directed towards learning frequency distributions of (discrete) values held by clients, such as commonly visited web sites.  Since rare values are likely to be privacy-revealing (e.g., URLs that encode identifying information), we seek  popular values, or ``heavy hitters.''  
Google researchers show that sampling a random (secret) subset of clients can achieve differential privacy~\cite{fedhh}.  
PRIO also targets towards secure computation of heavy hitters, where DP noise can be introduced by the aggregating servers~\cite{PRIO, DPRIO}.  
Other approaches 
are via distributed noise addition, where each client adds small random noise~\cite{Bagdasaryan:23}, or via shuffling  client responses to disassociate the message from the sender, which has a similar statistical effect~\cite{Prochlo, talwar:23}.  
Explicitly suppressing responses with low frequencies captures the intuitive privacy definition of $k$-anonymity~\cite{Star}. 

\section{Discussion and Concluding Remarks}
We have laid out our experience and insights in building a state-of-the-art large scale FA system.
\redit{
The chief takeaways that we draw from our experience are: 
(1) It is indeed possible to achieve large scale, secure federated analytics.  At this scale, failures are inevitable, and so it was important for us to design with these in mind (e.g, incorporating snapshotting and recovery). 
(2) Security is increasingly available as a commodity, thanks to availability of secure hardware.  Designing for simplicity ensured that security was auditable and efficient for us.
(3) A wide range of tasks are served by the Secure Sum and threshold operation, including finding popular values, capturing distributions, and gathering feature usage statistics. 
(4) We obtained privacy in addition to security by designing in noise addition (differential privacy), and at scale this made minimal impact on accuracy.}

\edit{
Key design choices that could be revisited in future are:}

\smallskip
\paragraph{Privacy Models and TEEs.}
\edit{
The FA system accommodates various privacy models (central, local and distributed DP).  
The use of TEEs is of particular importance in allowing complex central DP algorithms, and the secure aggregation for local and distributed DP. 
Current TEE hardware comes with some tradeoffs, which will alter as new generations are released. 
Our threat model treats data poisoning efforts  as negligible, but protections against malicious clients could be hardened.}

\smallskip
\paragraph{Privacy Budgeting.}
\edit{
DP parameter $\epsilon$ captures the degree of privacy, which is weakened as more queries are answered.  
Our pragmatic approach is to focus on per-query privacy impact, and seek to ensure that analysts do not posed repeated queries to the same data. 
In future we may use stronger constraints, and more sophisticated privacy accounting tools~\cite{elephantdp,DoroshenkoGKKM22}.}

\smallskip\paragraph{Longitudinal Issues.} 
\edit{
Currently, we allow clients to respond to one batch of queries per day, pertaining to data collected over the previous 24 hours.  
We will loosen these constraints in future versions as we allow more fine-grain responses.} 

\smallskip
\paragraph{Other considerations.}
\edit{When we try to batch queries together for efficiency, we can combine responses while ensuring that the results do not expose too much private information. 
For analytics, the main demand is for aggregation queries, future plans are to expand to more types of queries and data. 
As we extend the deployment of our FA stack to more use cases, we will face new types of queries, and have to decide whether to answer these via custom code, or extend the capability of our system to provide more ``built-in'' support for common query patterns.  
A leading concern is how to simplify the deployment of queries by non-expert analysts, and in particular how to provide useful debugging and cross-checking information, while maintaining the high levels of privacy guarantees.
}

\section*{Acknowledgements}
Contributions to this project extend beyond the authors. We are grateful for the following people who have engaged in system design/implementation and provided valuable feedback: Ilya Mironov, Jon Millican, Eric Northup, Ravi Ranjan, Anthony Shoumikhin, Pavel Ustinov, Renchang Miao, Vlad Grytsun, Igor L. Markov and Akash Bharadwaj.

\clearpage
\appendix

\section{Federated Algorithms: Quantiles}
\label{sec:algo}
The federated model opens itself to a number of design techniques to answer a wide range of queries. 
Simple queries, such as sums and counts, are straightforward to implement within FA,
while more complex queries require more nuanced solutions. 
Examples include
finding the ``heavy hitters'' (frequent items) from a large domain~\cite{fedhh}, 
estimating U-statistics across pairs of observations~\cite{BellBGK20}, 
finding the mean of scalar and vector values~\cite{VargaftikBPMBM21},
measuring classifier accuracy and AUC~\cite{Cormode:Markov:23}, 
and
making heat maps of data distributions~\cite{Bagdasaryan:23}. 

For FA algorithms, there is a strong preference for techniques that require only one or a very few (constant) rounds of data collection. 
This is distinct from Federated Learning, where it is common to iterate towards a solution over hundreds or thousands of short-lived rounds. 
It is also highly desirable to express the algorithms using well-supported tooling, specifically the SST primitive. 
We favor algorithms that are robust to clients drop outs and unavailability, via sampling and ensuring that client information is encapsulated in a single message. 

To illustrate the design space and considerations of building support for a non-trivial query into a federated analytics system, we study an example query in depth: to find the quantiles of a (distributed) data distribution. 
\label{sec:quantiles}
Finding quantiles (or percentiles) is one of the most requested queries for the FA stack.  
We have made several implementations to handle different trust and privacy models, and thus refined our understanding and optimized the approach that we advocate. 
Our first efforts used multiple rounds of interaction, which slowed down the process, and led to synchronization issues.  
Subsequently, we focused on approaches using few rounds. 


Given a (multi)set of readings, the quantile query seeks the point $p$ such that a $q$ fraction of readings are below it.  
For instance, the median asks for the 0.5-quantile, where half the readings are below and half are above.
We also often seek the 90\%-ile, 95\%-ile and the 99\%-ile 
corresponding to the 0.9-, 0.95- and 0.99-quantiles.  
There is noise in the reported answer as we use approximate algorithms and introduce noise for (differential) privacy reasons. 
%
%
Outside of FA, there are algorithms to build compact summaries for the quantiles problem, particularly when processing the inputs as a single stream of data. 
These include the q-digest~\cite{qdigest}, t-digest\cite{tdigest}, dd-sketch~\cite{ddsketch} and Greenwald-Khanna (GK) summary~\cite{Greenwald:Khanna:01}.  
However, these do not all immediately map to the federated setting, nor do they provide a privacy guarantee.  

The simplest approach to answering a fixed quantile query in the federated setting is to perform a binary search over multiple rounds.
We start with a range $[\texttt{low}, \texttt{high}]$ that all the data falls in, and issue a federated counting query to find what fraction of examples fall in this range.  
This process is repeated
for ranges $[\texttt{low}, p]$, adjusting $p$ higher and lower based on the result of the previous rounds, until we find an answer for which the count is close enough to the target $q$ (as a fraction of the number of participating clients).  
Typically, 8-12 rounds suffice, provided the initial range is fairly tight around the true data. 
However, this can be  slow to complete. 

The key to reducing the number of rounds is to think of
the multi-round algorithm 
as looking up values in a collection of histograms of progressively finer granularity.  
The first round looks up the two values in a two bucket histogram, where the buckets correspond to the left and right halves of the value domain, respectively.  
Depending on what answer is obtained, the second round looks up two values in a four bucket histogram (corresponding to dividing the domain into four equal pieces), and so on.  
Importantly, although the set of buckets to inspect depends on what we find in the previous round, the choice of bucket boundaries is not data dependent.  
Hence, we can build out the complete set of histograms in a single round of FA, and use the output of this query to answer all-quantiles queries. 
We refer to this as the hierarchical, or \textit{tree}, approach to quantile estimation. 

To put this into practice
we need some prior knowledge of a range that all values fall into (or a first round of FA to gather this knowledge), in order to specify the histograms.  
We must determine the number of histograms to build,
based on the granularity of the finest histogram.  
Building histograms out to a depth of 12 (giving $2^{12}$ buckets at the finest level of detail) gives a good level of accuracy in practice, while remaining relatively quite compact. 
In some cases, rather than the full hierarchy, it suffices to collect data at the finest level and treat the resulting (noisy) histogram as if it gave the exact distribution. 
We refer to this as the flat, or \textit{hist}, approach. 


\subsection{Experiments on Quantile Data Collection}

\begin{figure*}[t]
\ifthenelse{\tech}{
\centering
    \begin{subfigure}{0.45\textwidth}
        \centering
        \includegraphics[width=1\textwidth]{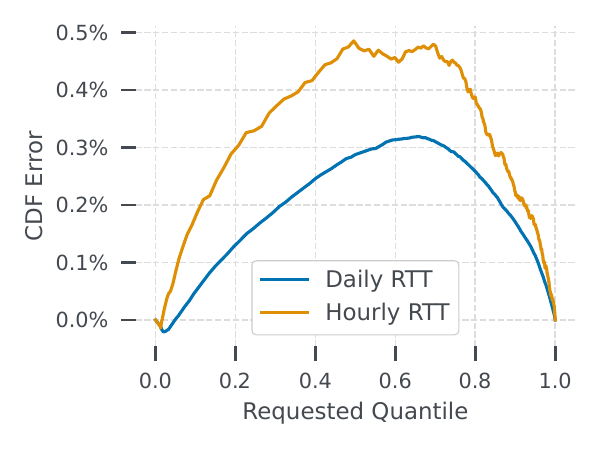}
        \caption{CDF approximation error for RTT}
        \label{fig:cdf-error}
    \end{subfigure}%
    \begin{subfigure}{0.45\textwidth}
        \centering
        \includegraphics[width=1\textwidth]{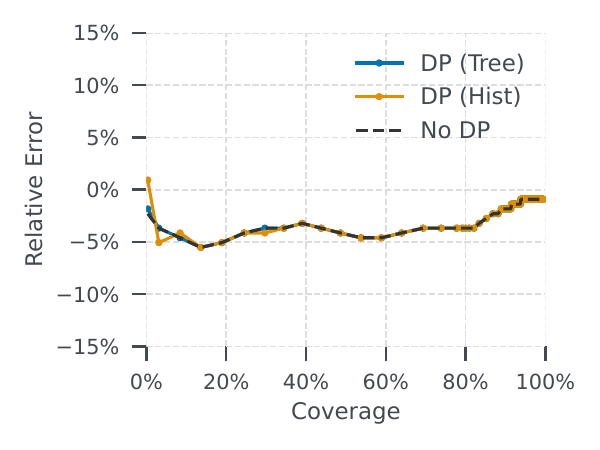}
        \caption{DP error for daily 90\%-ile daily RTT}
        \label{fig:rtt-quantile}
    \end{subfigure}
    \begin{subfigure}{0.45\textwidth}
        \centering
        \includegraphics[width=1\textwidth]{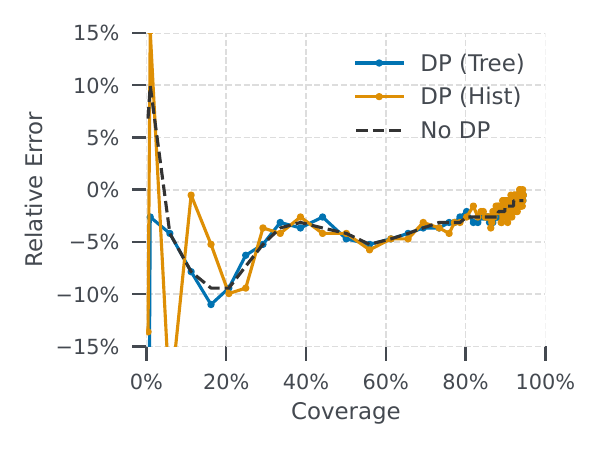}
        \caption{DP error for 90\%-ile hourly RTT} 
        \label{fig:hourly-quantile}
    \end{subfigure}
}{ \centering
    \begin{subfigure}{0.33\textwidth}
        \centering
        \includegraphics[width=1\textwidth]{plots/cdf_error_combined.pdf}
        \caption{CDF approximation error for RTT}
        \label{fig:cdf-error}
    \end{subfigure}%
    \begin{subfigure}{0.33\textwidth}
        \centering
        \includegraphics[width=1\textwidth]{plots/linear_rtt-quantile.pdf}
        \caption{DP error for daily 90\%-ile daily RTT}
        \label{fig:rtt-quantile}
    \end{subfigure}%
    \begin{subfigure}{0.33\textwidth}
        \centering
        \includegraphics[width=1\textwidth]{plots/linear_hourly_rtt-quantile.pdf}
        \caption{DP error for 90\%-ile hourly RTT} 
        \label{fig:hourly-quantile}
    \end{subfigure}
}
    \caption{CDF error after 48 hours \textit{(left)} and relative error in DP estimation of 90th percentile RTT \textit{(center, right)}}
    \label{fig:quantile}
\end{figure*}


\edit{
We implement and apply these quantile algorithms in order to evaluate the impact of our design choices on query accuracy and scalability. }
Figure~\ref{fig:cdf-error} shows the error in estimating the cumulative distribution function (CDF) of the number of data points per device
, based on histograms of $B = 2048$ buckets representing sampled counts of $1, 2, ..., 2047, 2048+$.
That is, for each potential quantile query (e.g., the median, being the 0.5-quantile), we identify which true quantile the reported value corresponds to, using knowledge of the ground truth distribution. 
We plot this for both daily and hourly round-trip times, using data collected after 48 hours from initiating the query. 
The pattern for both is similar: the error is zero at either extreme, since the measure  can be satisfied by reporting an arbitrarily small value for the 0.0-quantile, and an arbitrarily large value for the 1.0-quantile. 
We gain more information when studying quantiles that are closer to the middle of the distribution. 
After 48 hours of data collection, the maximum error is 0.32\% for the daily measurements, and 0.49\% for the hourly data---this is the Kolmogorov-Smirnov test statistic for measuring similarity of distributions.  
In both cases, it is much less than 1\%, indicating close agreement. 
The error is higher in the hourly case, due to fewer observations.

Figures~\ref{fig:rtt-quantile}, \ref{fig:hourly-quantile} show our results for estimating the 90th percentile (i.e., 0.9-quantile), under different privacy approaches.
Per Appendix~\ref{sec:quantiles}, we add noise either to a flat histogram, \textit{DP (hist)}, or to a hierarchy of histograms, \textit{DP (tree)}. In both cases, we follow a central DP model, adding Gaussian noise to the histograms to achieve an $(\epsilon, \delta)$-DP guarantee for $\epsilon=1, \delta=10^{-8}$.
We compare the accuracy of these quantile estimates against a federated quantile estimate without any privacy noise addition (\textit{No DP}). 
Here, we measure the relative error: we compute the ratio of reported value to the ground truth value of the 90th percentile.  

Unsurprisingly, when only a few clients have reported, there is a lot of uncertainty in the value of the target quantile, particularly for the hourly data (Figure~\ref{fig:hourly-quantile}).  
However, once more than about a quarter of clients have submitted their input, we can find a reliable estimate of the quantile, up to a few percentage points in variation. 
For these experiments, where we have many clients each reporting a single contribution to the histogram, the impact of DP noise is marginal: the greater uncertainty comes from the sampling effect of partial client participation. 
The tree histogram method adheres closer to the no-DP case than the simple flat histogram.
Our practical experience is that the tree approach is  preferred when larger histograms are needed to represent the input.
We conclude
that we can obtain very accurate answers to quantile queries while enjoying FA's strong security and privacy properties.



\newpage
\balance
\bibliographystyle{abbrv}
\bibliography{fa}

\end{document}